%% file: main.tex
\pdfoutput=1

\documentclass[11pt]{article}

\usepackage[final]{acl}

\usepackage{times}
\usepackage{newtxmath}
\usepackage{latexsym}

\usepackage[T1]{fontenc}

\usepackage[utf8]{inputenc}

\usepackage{microtype}

\usepackage{inconsolata}

\usepackage{graphicx}

\usepackage{booktabs}
\usepackage{tabularx}
\usepackage{multicol}
\usepackage{multirow}
\usepackage{soul}
\usepackage{caption}
\usepackage{subcaption}
\usepackage{listings}
\usepackage[frozencache=true,cachedir=minted-cache]{minted}

\definecolor{LightGray}{gray}{0.9}

\title{When and Where Did it Happen?\\ An Encoder-Decoder Model to Identify Scenario Context}

\author{
 \textbf{Enrique Noriega-Atala\textsuperscript{1}},
 \textbf{Robert Vacareanu\textsuperscript{1}},
 \textbf{Salena Torres Ashton\textsuperscript{2}},
\\
 \textbf{Adarsh Pyarelal\textsuperscript{2}},
 \textbf{Clayton T. Morrison\textsuperscript{2}},
 \textbf{Mihai Surdeanu\textsuperscript{1}}
\\
\\
 \textsuperscript{1}Department of Computer Science,
 \textsuperscript{2}College of Information Science\\
 The University of Arizona
\\
 \small{
  \texttt{\{enoriega,rvacareanu,salena,adarsh,claytonm,msurdeanu\}@arizona.edu}
 }
}

\begin{document}
\maketitle
\begin{abstract}
We introduce a neural architecture finetuned for the task of \emph{scenario context} generation: The relevant location and time of an event or entity mentioned in text. Contextualizing information extraction helps to scope the validity of automated finings when aggregating them as knowledge graphs. Our approach uses a high-quality curated dataset of time and location annotations in a corpus of epidemiology papers to train an encoder-decoder architecture. We also explored the use of data augmentation techniques during training. Our findings suggest that a relatively small fine-tuned encoder-decoder model performs better than out-of-the-box LLMs and semantic role labeling parsers to accurate predict the relevant scenario information of a particular entity or event.
\end{abstract}

\input{sections/introduction}
\input{sections/related_work}
\input{sections/dataset}
\input{sections/experiments_and_results}
\input{sections/conclusions_and_future_work}
\input{sections/limitations_and_ethical_considerations}
\input{sections/acknowledgments}
\bibliography{anthology, custom}

\appendix
\input{sections/annotations}
\input{sections/model_inputs_and_outputs}
\input{sections/baseline_prompt}

\input{sections/data_augmentation}

\end{document}

%% file: sections/introduction.tex
\section{Introduction}
We present an approach to contextualizing information extraction (IE) that
focuses on enhancing events and entities with \emph{scenario context}: the location and time relevant to extracted elements.

Knowing \emph{when} and \emph{where} an event occurs has become increasingly relevant due to the wide adoption of large-scale machine reading technology.
Decision makers in high-stakes areas,\footnote{\scriptsize
\url{https://www.darpa.mil/program/automating-scientific-knowledge-extraction-and-modeling}}
like epidemiology, public health or climate sciences, are increasingly turning
to natural language processing (NLP) technologies to help guide their decision making through
automatic evidence discovery and aggregation. In light of this, properly
scoping automated IE becomes very valuable to the users of these tools.

One example of a domain when scenario context information is relevant is the
modeling of epidemic dynamics, where the literature describes different
outbreaks using different mathematical models, such as variations of the
susceptible-infected-recovered (SIR) compartmental model. The different
scenarios have different parameters and it is useful to contextualize the
relevant event to have a better picture of the scenario described.
Another example is the domain of climate and climate change, where changes
in the climate of different geographical regions over time is studied by the
geosciences community.

Scenario information is often explicitly found in the periphery of the text describing an extraction, but not necessarily
in the same sentence---thus, it is a form of \emph{inter-sentence} relation
extraction (see \autoref{fig:examples} for examples).

In this work, we tackle the problem as a \emph{generative} task using an encoder-decoder transformer based on T5~\cite{Raffel2019ExploringTL}.
Given the locations and temporal phrases in an input passage, we prompt the
model to chose and generate the relevant scenario information with respect to a
specific entity or event.
The main contributions of this work are the following:

\noindent\textbf{(1)} An encoder-decoder model finetuned for generating \emph{scenario
context}, i.e., the spatial and temporal context of a particular event or concept within a larger phrase.

\noindent\textbf{(2)} A high-quality, hand-curated dataset of location and temporal
relations with intra- and inter-sentence relations, used to train and evaluate
the aforementioned model.

\noindent\textbf{(3)} An error analysis of the predictions of the model,
shedding light on potential future improvements to this method.

All artifacts used to train and evaluate the model
are publicly available.\footnote{Code, data and artifacts available for
download from \url{https://github.com/ml4ai/scenario-context}.}

%% file: sections/related_work.tex
\section{Related Work}
\begin{figure*}[t]
	\centering
	\begin{subfigure}[b]{1.8\columnwidth}
	  \includegraphics[width=\columnwidth]{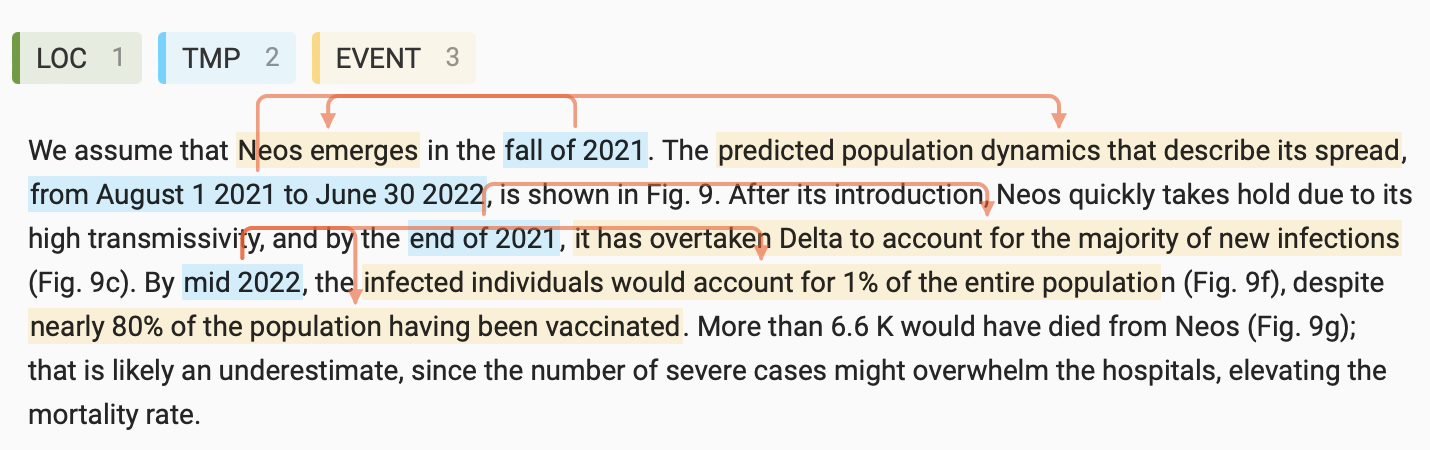}
	  \caption{Temporal annotations}\label{fig:temporal}
	\end{subfigure}
	\hfill\\
	\begin{subfigure}[b]{1.8\columnwidth}
	  \includegraphics[width=\columnwidth]{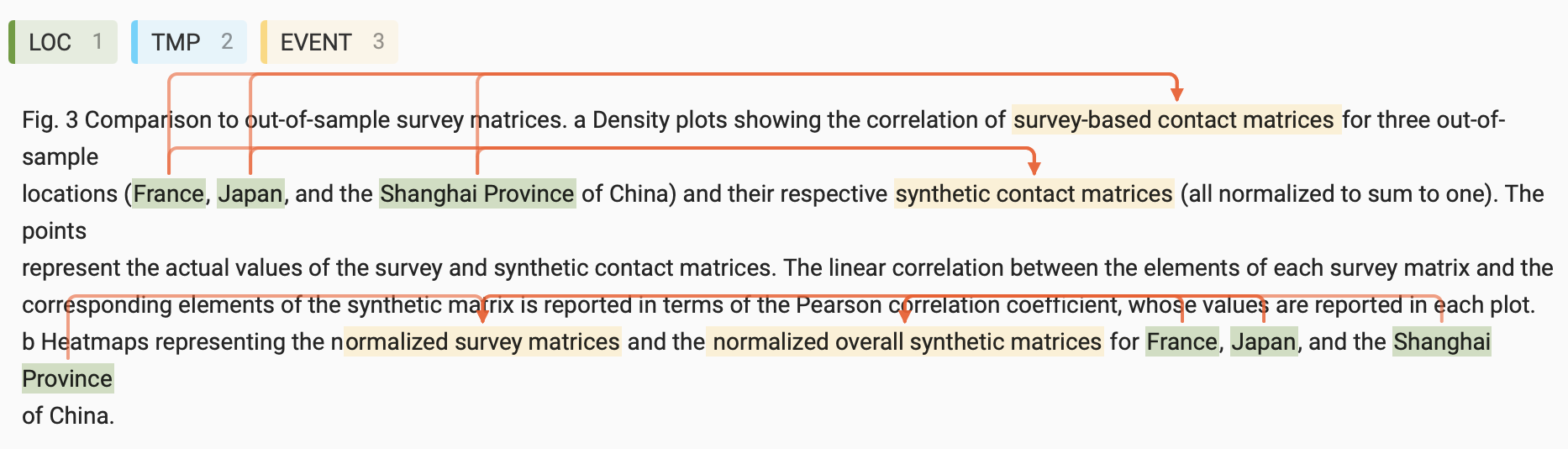}
	  \caption{Location annotations}\label{fig:spatial}
	\end{subfigure}
	\caption{%
        Example annotations in our dataset. Predicates
        highlighted in \textcolor{Dandelion}{yellow} represent `events' with scenario context
        information assigned to them, text highlighted with
        \textcolor{Cyan}{cyan} represents temporal context,
        and text in \textcolor{Green}{green} location context. The arrows connect a context
        expression to an event they are associated with. The scenario context
        to event associations are effectively many-to-many relations.
        \autoref{fig:temporal} shows a passage with several temporal scenario
        contexts
        and \autoref{fig:spatial} shows several location scenario contexts.
}
  \label{fig:examples}
  \end{figure*}
Annotating \textit{when} and \textit{where} an event occurs is closely related
to semantic role labeling (SRL)~\cite{levin2005argument, Gardner2017AllenNLP}
and document-level relation extraction~\cite{sahu-etal-2019-inter, xu-etal-2022-document, Delaunay2023ACS}.
An important distinction, however, is that SRL operates on parsing the
structure of a sentence and assign roles to phrases within it. Our proposed
approach is generative in nature, and is not constrained to the structural
elements identified during parsing.

The typical approach for IE
is finetuning a pre-trained transformer~\cite{Vaswani2017AttentionIA} such as T5 on the data of interest.
We build upon the family of encoder-decoder architectures~\cite{Bahdanau2014NeuralMT} to map an input passage to the expected output that represents the relevant scenario information.

Recently,
large language models (LLMs) \cite{brown2020language, jiang2023mistral,
Touvron2023LLaMAOA} have gained traction, but concerns remain about their
tendency to hallucinate.
Our work relies on supervised learning via finetuning instead of attempting to mitigate hallucinations.

Contextualizing IE has been of interest to the research community for a while, particularly in the biomedical domain~\cite{NoriegaAtala2018InterSentenceRE,Sosa2022ContextsAC,DBLP:journals/corr/abs-2112-09288}.
In this work we focus instead on the more general class of \emph{scenario information}: the relevant location and time of an entity or event.

%% file: sections/dataset.tex
\section{Dataset}

In order to train our model, we created a dataset that contains location and
temporal context annotations at both intra- and inter-sentential levels.
We focused on 22 epidemiology research articles, including ones that involve
modeling the dynamics of infection and outbreak case-studies, published between
2020--2022.

These articles often describe parallel scenarios to compare
and contrast the behavior of different outbreaks, making
the inference of the scenario context of relevant concepts and events in
these papers non-trivial.
Correctly understanding the location and time period for specific
events is important for the accuracy of any inference drawn from these studies.
We excluded any temporal mentions that were abstract or relative, and any
location mentions that were modifiers or adjectives. \autoref{fig:examples}
shows an example of an annotated passage of each kind.

The dataset comprises 383 passages, ranging in length from a single
sentence to a couple of paragraphs. A total of 1,382 relations were annotated with
833 (60.3\%) location context relations and 549 (39.7\%) temporal context
relations.

\begin{figure}[!htb]
  \centering
  \includegraphics[width=\columnwidth]{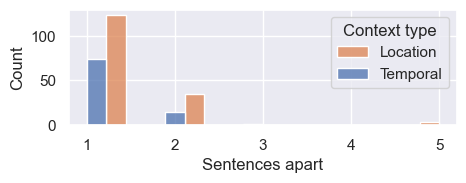}
  \caption{Number of sentences between the relevant entity/event and its corresponding context.}
  \label{fig:distances}
\end{figure}

A key aspect of the dataset is the presence of annotations for inter-sentential
relations---i.e., relations where the event/concept of interest and its
scenario context are in different sentences. These comprise 18\% of scenario
context relations.
Approaches that rely on syntax, e.g., SRL, are not
well suited for inter-sentential relations. \autoref{fig:distances} shows the distribution of sentence distances for the inter-sentential subset of annotations.

A random sample consisting of 13\% of the relations was used to measure
inter-annotator agreement using the Cohen's Kappa
method~\cite{cohen1960coefficient}, resulting in a score of
$\kappa=0.79$.\footnote{Agreements over 0.61 are considered substantial and
over 0.81 are considered almost perfect~\cite{Landis1977TheMO}.}. Details about the annotation guidelines are found in \autoref{appendix:annotations}

\subsection{Data Augmentation}

Manually annotating data is time-consuming and labor-intensive. To address this
challenge and scale up the amount of data available for the scenario context
generation task, we explored two techniques for data augmentation using
LLMs---paraphrasing and procedural generation.
The details of prompts and generation procedures are provided in \autoref{appendix:dataaug}.

\noindent \textbf{Paraphrasing}. To increase the lexical and syntactic diversity of the gold annotations, we generated variations of each passage in the dataset using GPT-4.
Additionally, we substituted the temporal and location arguments with alternatives while keeping track of the relations present in them. 
This process resulted in 434 additional scenario context relations.

\noindent\textbf{Procedurally generated relations}. We used GPT-4 to
procedurally generate passages containing one or more fictional events with temporal and location context.
We used the prompt to control the topic, role of the narrator, length, and number of scenario contexts in each passage.
This procedure allows to scale up the amount of training data proportional to
one's budget.
This approach resulted in an additional 1,361 scenario context relations.


%% file: sections/experiments_and_results.tex
\section{Experiments and Results}\label{sec:experiments}

\begin{table*}
  \centering
  \footnotesize
  \begin{tabular}{lrrrrrrrrrrrr}
    \toprule
    &\multicolumn{6}{c}{Span-level} & \multicolumn{6}{c}{Token-level}\\
      \cmidrule(lr){2-7}
      \cmidrule(lr){8-13}
    &
      \multicolumn{3}{c}{Location} &
      \multicolumn{3}{c}{Temporal} &
      \multicolumn{3}{c}{Location} &
      \multicolumn{3}{c}{Temporal} \\
      \cmidrule(lr){2-4}
      \cmidrule(lr){5-7}
      \cmidrule(lr){8-10}
      \cmidrule(lr){11-13}
      Model & P & R & F1 & P & R & F1  & P & R & F1 & P & R & F1 \\
    \midrule
    Annotations & 	\textbf{0.81}	& \textbf{0.80} &	\textbf{0.80}	& 0.76 &	0.73 &	0.74 &	\textbf{0.84}	& \textbf{0.84}	& \textbf{0.83}	& 0.76	& 0.80	& 0.77\\
    ~~+paraphrases &	0.80	& 0.79 &	0.78 &	0.78 &	0.76 &	0.76 &	0.83	& 0.82	& 0.82	& \textbf{0.79}	& \textbf{0.81}	& \textbf{0.79}\\
    ~~+synthetic	&	0.78	& 0.77 &	0.76 &	\textbf{0.79} &	\textbf{0.78} &	\textbf{0.78} & 0.81
                    & 0.81	& 0.80	& \textbf{0.79}	& \textbf{0.81} &	\textbf{0.79}\\
    ~~+para \& synth	&	0.77	& 0.76	& 0.75	& 0.71	& 0.70	& 0.7 &0.81
                        &	0.81	& 0.80	& 0.74	& 0.75	& 0.74\\
    \bottomrule
  \end{tabular}
  \caption{Scenario context evaluation results. P = Precision, R = Recall.}
  \label{tab:main-results}
\end{table*}


\begin{table}
  \centering
  \footnotesize
  \begin{tabularx}{\linewidth}{Xrrrrrr}
    \toprule
    &
      \multicolumn{3}{c}{Location} &
      \multicolumn{3}{c}{Temporal}
      \\
      \cmidrule(lr){2-4}
      \cmidrule(lr){5-7}
    Method     & P    & R    & F1   & P    & R    & F1  \\
    \midrule
    GPT-4o     & \textbf{0.29} & \textbf{1}    & \textbf{0.44} & \textbf{0.33} & 0.89 & \textbf{0.48}\\
    Mistral-7B & 0.25 & \textbf{1}    & 0.40 & 0.23 & \textbf{0.98} & 0.38\\
    \midrule
    SRL        & 0.09 & 0.01 & 0.02 & 0.08 & 0.05 & 0.06\\
    \bottomrule
  \end{tabularx}%
  \caption{SRL and LLM baseline results. P = Precision, R = Recall, Mistral-7B
  = Mistral 7B Instruct v0.2.}\label{tab:srl-baseline}
\end{table}

We trained\footnote{The model was finetuned using a workstation with an RTX
3090Ti GPU, a Threadripper 3960X 24-Core CPU and 128 GB of system memory. Each
model was finetuned from \texttt{t5-base} using HuggingFace's
\texttt{Seq2SeqTrainer} for 10,000 steps, 3e-5 learning rate, linear weight decay of 0.1
and batch size of 4.} an encoder-decoder model~\cite{Sutskever2014SequenceTS, Vaswani2017AttentionIA} based on T5
to generate the location and temporal information relevant to a
specific event from its surrounding context.
For each relation in the dataset, we prompted (see \autoref{appendix:prompt}
for details)
the model to decode the context information of the specific event.
Each event may have zero, one or more context relations of each type. The model
decoded all of them simultaneously.

We held out a random sample of 20\% of the annotions for testing and
fine-tuning
\texttt{t5-base}.
\autoref{tab:main-results} contains the main results averaged across three runs
with different random seeds.
Since a particular event may have zero or more annotations of each type, we compute precision, recall, and F1 individually for each and average them across the testing set.
We report two variants of this evaluation: (i) span-level, and (ii) token-level.
At the \emph{span level}, a generation is considered correct only if it exactly matches the gold standard annotation.
In order to ignore minor lexical variations, we applied a basic normalization procedure before comparing strings: converting to lowercase, trimming spaces on both ends, and removing commas. 
Nevertheless, having a partially correct prediction may still be useful (e.g., \texttt{july 5 1987} vs \texttt{july 1987}), therefore the \emph{token level} evaluation reports the precision, recall, and F1 scores at the token level, similar to SQuAD \cite{Rajpurkar2016SQuAD1Q}.

\autoref{tab:main-results} contains the results of models trained (i) with
manual annotations only and (ii) with the augmented dataset.
In general, our approach achieves better results when predicting location than temporal context.
Training only with the manual annotations results on the best performance for
locations; training with the augmented data decreases the performance for location,
but improves the performance for temporal context.
We hypothesize that this is a consequence of the differences between expressing
named locations vs. temporal expressions. There is less variance in how
locations are written; they are usually proper nouns or adjectives, whereas
time expressions are much more varied---
they could be expressed as a standalone year, full date, date range, season, relative temporal phrase, etc.
Data augmentation may help increase the diversity of temporal context phrases shown during training, leading to less overfitting to the lexicon compared to location.

\paragraph{Baselines}

We compare our methods with a decoder-based LLM approach and an SRL system.
\autoref{tab:srl-baseline} contains the baselines' results.
For the LLM baseline, we tested GPT-4o~\cite{openai2023gpt4} and Mistral 7B~\cite{jiang2023mistral}.
We asked the models\footnote{Prompting details in \autoref{appendix:llmbaseline}} to generate the scenario context for each event and computed the span level results.
We find that the LLMs successfully identify time spans and locations relevant to concepts and events, but also tend to predict spurious relations that are not related to the focus of the query.
This is reflected in the high recall and low precision exhibited by the LLMs.
These observations support the use of supervised learning approaches when feasible.

SRL assigns roles between the clauses in a sentence.
We used it as an alternative baseline to the other generative approaches.
To test for scenario context detection, we used AllenNLP's structured prediction pre-trained model\footnote{\url{https://storage.googleapis.com/allennlp-public-models/structured-prediction-srl-bert.2020.12.15.tar.gz}} to parse the sentences containing events in each passage of the ground truth dataset.
We considered a scenario context relation `extracted' if the appropriate context is contained within a predicted modifier argument (\texttt{ARGM-LOC}, \texttt{ARGM-TMP}) and the text of the event is contained in the union of another argument with the predicate.
We found that SRL is not well-suited for this task---it often failed to select the event of focus within an argument.

\paragraph{Error Analysis}

We performed an error analysis on a sample of the testing predictions of the model trained only with human annotations.
\autoref{tab:error} contains different types of prediction errors broken down by scenario context type.
\textit{Spurious predictions} occur when there is no context annotation, but
the model generates a prediction; conversely, a \textit{Missing prediction} happens when there is a gold annotation but no prediction from the model.
\textit{Mistaken predictions} are when there is both a gold annotation and a prediction, but the model was outright wrong about it.
\textit{Partial predictions} occur when the generated text is properly
contained in the annotation's text, but is not an exact match---e.g., an event with a location context annotation of \texttt{``Western and Northern Europe, United Kingdom''} where the model predicted \texttt{``Western and Northern Europe''} is a partial prediction; \textit{Overprediction} errors are the opposite.
These instances are considered false positives for the span-level results in
\autoref{tab:main-results}, however their partial, accurate predictions are
accounted for in the token-level evaluations.
\emph{Other} errors are artifacts of the generative nature of the task.
Consider the gold annotation \texttt{``California, Indiana, New York''} and the
prediction of \texttt{``California (CA), Indiana (IN), New York (NY)''}---clearly the prediction is correct; however, the model decoded state acronyms alongside the names, which fail an exact string match.
The token-level evaluation is able to pick up the full state names.
For temporal context, \emph{Mistaken} predictions mostly stem from the variety
of ways time intervals can be expressed in text and the inability of the model
to abstract that information, e.g., the gold annotation \texttt{``between 2009 and 2014''} was predicted as \texttt{``2009, 2014''}, which correctly includes the endpoints of the range, but fails to specify the crucial detail that this is an inclusive range including the years between them.



\begin{table}
  \centering
  \footnotesize
  \begin{tabular}{lrr}
    \toprule
    & \multicolumn{2}{c}{Context Type}
    \\
      \cmidrule(lr){2-3}
    Error Type & Location & Temporal\\
    \midrule
    Spurious   & 11       & 12 \\
    Missing    & 10       & 7 \\
    Mistaken   & 4        & 6\\
    \midrule
    Partial    & 7        & 3 \\
    Over       & 8        & 7 \\
    Other      & 3        & 4 \\
    \midrule
    Total      & 49       & 39\\
    \bottomrule
  \end{tabular}%
  \caption{Prediction error types.}\label{tab:error}
\end{table}

%% file: sections/conclusions_and_future_work.tex
\section{Conclusions and Future Work}

In this work, we introduced an encoder-decoder model finetuned to generate
location and temporal context associated with a particular concept or event. We
are releasing a dataset of hand-curated annotations from a collection of
academic papers in the epidemiology domain that describe the dynamics of
outbreaks in different locations and times. We found that our method
more accurately recognizes the relevant context than out-of-the-box LLMs
or SRL. We also explored the use of data augmentation
methods, finding that they resulted in modest improvements in temporal context
extraction.

There are at least two promising avenues for future work. The first is
expanding the dataset to include more curated annotations from additional
domains.  This will foster the development of more accurate models with better
generalization capabilities.  The second is exploring other network
architectures, such as span-prediction or decoder-only models.  The former is
useful for attributing the source of the context prediction and the latter can
benefit from the transfer learning potential exhibited by open-source LLMs.

%% file: sections/limitations_and_ethical_considerations.tex
\section{Limitations and Ethical Considerations}
While the methods described in the paper are not specific to a particular
domain, the annotations focus on scientific literature in the domain of
epidemiology. The evaluations carried out in this work did not test for
generalization capabilities on different domains. Additionally, all of the
information used in this work was written solely in English, limiting the potential impact and applications of our contributions.
While we evaluated the performance of LLMs for this task, we only tested two different models: GPT-4o and Mistral-Instruct. We recognize that the landscape of LLMs changes quickly and that the state-of-the-art is fleeting. Due to this, our baseline results may be rendered obsolete in the near future.

%% file: sections/acknowledgments.tex
\makeatletter
\@ifpackagewith{acl}{review}{}{
\section*{Acknowledgments}

Research was sponsored by the Defense Advanced Research Projects Agency and
was accomplished under contract number HR00112290092.

The views and conclusions contained in this document are those of the
authors and should not be interpreted as representing the official
policies, either expressed or implied, of the U.S. Government.
}
\makeatother

%% file: sections/annotations.tex
\section{Annotation Guidelines}\label{appendix:annotations}
We used LabelStudio\footnote{\url{https://labelstud.io}} to manually annotate scientific articles with scenario context information. LabelStudio was set up with 383 \emph{tasks}, where each contains a section of the article's text containing either location or temporal scenario information of a specific event.
At least one annotator carefully read each passage, selecting all the events
with a designated location and/or temporal information. The annotator then
proceeded to select and link each piece of scenario context information to the
relevant event. \autoref{fig:examples} displays two examples of the user
interface of LabelStudio with different types of scenario context and event information.

Two other independent annotators worked in a sample comprising 13\% of the tasks. Using
these additional annotations, we computed an inter-annotator agreement metric using Cohen's Kappa of $\kappa = 0.79$.

%% file: sections/model_inputs_and_outputs.tex
\section{Model Inputs and Outputs}
\label{appendix:prompt}
\autoref{lst:input} shows the format of the prompt used as input to the scenario context model.
Fields between double curly braces are substituted with the text containing the entity or event of focus in \texttt{\{\{event\}\}} and the complete passage from which relevant scenario context will be retrieved in \texttt{\{\{context\}\}}.

\autoref{lst:output} shows the output format produced by the model.
The model will generate zero or more relevant locations and time expressions per input.
Double curly braces are placeholders for the actual predicted values decoded by the model.

\begin{figure}[h]
  \centering
  \begin{verbatim}
    Text: {{event}}

    Context: {{passage}}
  \end{verbatim}
  \caption{Input prompt format used by the scenario context encoder-decoder model.}\label{lst:input}
\end{figure}

\begin{figure}[h]
  \centering
  \begin{verbatim}
    location: {{loc 1}}, ..., {{loc n}};
    time: {{tmp 1}}, ..., {{tmp n}}
  \end{verbatim}
  \caption{Output sequence format decoded by the scenario context encoder-decoder model.}\label{lst:output}
\end{figure}

%% file: sections/baseline_prompt.tex
\section{LLM Baseline Prompt Template}
\label{appendix:llmbaseline}

The prompt template shown in \autoref{fig:baselineprompt} was to generate the scenario context predictions from both LLMs in \autoref{sec:experiments}. At run-time, \texttt{\{\{event\}\}} was substituted by the entity/event of focus and \texttt{\{\{pre\_context\}\}} and \texttt{\{\{post\_context\}\}} were subsituted by the passage's text before and after it, respectively.

The output of the LLMs was parsed as a \texttt{JSON} object and used to compute the baseline scores.

\begin{figure}[h]
	\centering

	\begin{verbatim}
For the following phrase, look at the 
event or concept surrounded by ``` and 
tell me the locations and time periods 
that relevant to the element surrounded
by ```.
The output format should be a json
object with an array of strings for 
type of context. If there is not any 
element of a specific type, you will 
put an empty array in its value.
Output format:
	{
		"locations": [],
		"time periods": []
	}
Phrase:
{pre_context}```{event}```{post_context}
\end{verbatim}
	\caption{Prompt used to elicit scenario context using an LLM.}\label{fig:baselineprompt}
\end{figure}

%% file: sections/data_augmentation.tex
\section{Data Augmentation Procedures}
\label{appendix:dataaug}

\subsection{Paraphrasing Annotations}
We used GPT-4 to generate paraphrases of the annotated dataset. Each passage
was used as a seed to geenrate multiple paraphrases using the prompts listed in \autoref{fig:paraph}.

\begin{figure}[h]

\centering
\begin{verbatim}
- Please give me a location that is either
close or similar in nature with:
`{location}`.
Please do not return any additional
information.

- Please give me a date that is either close
or similar in nature with: `{date}`.
Please do not return any additional
information.

- Please rephrase the following text, while
keeping the following the following phrase
fixed:`{phrase}`
and maintaining the overall message and
length

{text}

- Please rephrase the following text,
maintaining the overall message and
length

{text}

- Please replace word `{word}` and its
derivatives with the word `{replacement}`
and its appropriate derivatives the
following text:

{text}
\end{verbatim}
\caption{Prompts used for parapghrasing sequences in the original dataset}\label{fig:paraph}
\end{figure}

\subsection{Procedurally Generated Data}
We used GPT-4 to procedurally generate synthetic data.

First, we seed the procedure with a set of event types. In our experiments we defined these to be \texttt{historical events}, \texttt{tech conferences}, and \texttt{public health emergencies}. Then, for each event type, we repeat the following steps:

\begin{enumerate}
	\item For each event type, we ask the LLM to generate ten different \emph{fictional} event names.
	\item We ask the LLM to generate five different \emph{narrator
        roles}---e.g., news reporter, high school student, historian, etc.
	\item We prompt the LLM to narrate each event, assuming each of the roles using a predefined set of numbers of paragraphs.
\end{enumerate}

\autoref{fig:procdata} contains a code snippet from a Jupyter notebook used to
generate the synthetic data.

\begin{figure*}[h]

	\centering
	\begin{minted}
		[
		framesep=2mm,
		xleftmargin=2em,
		fontsize=\footnotesize,
		linenos
		]
		{python}
# Get the number different types of events to develop context for

events_prompt = ChatPromptTemplate.from_messages([
	("user", "Generate a list of 10 different fictional {event_type} names.\
	Don't any include details, locations, names or dates. You must provide a comma-separated\
	list as a result.")
])

narrator_prompt = ChatPromptTemplate.from_messages([
	("user", "Generate a list of 5 different narrator roles that describe {event_type}.\
         For example, if we are describing a political event, a narrator role could be\
	 a news reporter; if it is a historical event, the narrator could be a historian\
	 writing a book, a highschool student writing a homework assignment or a PhD\
	 scholar writing a dissertation. You must provide a comma-separated list as the output.\
	 Don't include the event type, just the narrator role")
])

generation_prompt = ChatPromptTemplate.from_messages([
	("system", "You are a {role} describing {event_type}"),
	("user", """Write {length} about {event}.
	 Whenever you mention the event or refer, either explicitly or through pronoums, you
	 must wrap between <evt></evt> markup tags. You must include one location context in
	 your description. This location context represents where the event took place and
	 it could be a geographical region, country, city or location coherent with your
	 argument. Any mention of the geographical context must be  wrapped between
	 <loc></loc> markup tags. You must include {loc_distractors} other distractor
	 locations that are not related to the location context. It must be unambiguous, but
	 subtle, that these distractior locations are not where the event took place.
	 The distractor locations must be wrapped by <nloc></nloc> markup tags. You must
	 also include one temporal context in your description. This temporal context
	 represents the specific time or time frame in which the event took place. Any
	 mention of the temporal contect must be wrapped between <tmp></tmp> markup tags.
	 You must include {tmp_distractors} other distractor times that are not when the
	 event took place. It must be unambiguous, but subtle, that these distractor times
	 are not when the event happened. The distrator times be wrapped
	 by <ntmp></ntmp> markup tags.""".strip())
])

events_chain  = events_prompt | llm | list_parser
roles_chain  = narrator_prompt | llm | list_parser
generation_chain = generation_prompt | llm | str_parser
	\end{minted}
	\caption{Python code snippet with the prompts used to narrate a fictional event in order to procedurally generate data.}\label{fig:procdata}
	\end{figure*}

%% file: main.bbl
\begin{thebibliography}{19}
\providecommand{\natexlab}[1]{#1}

\bibitem[{Bahdanau et~al.(2014)Bahdanau, Cho, and
  Bengio}]{Bahdanau2014NeuralMT}
Dzmitry Bahdanau, Kyunghyun Cho, and Yoshua Bengio. 2014.
\newblock \href {https://api.semanticscholar.org/CorpusID:11212020} {Neural
  machine translation by jointly learning to align and translate}.
\newblock \emph{CoRR}, abs/1409.0473.

\bibitem[{Brown et~al.(2020)Brown, Mann, Ryder, Subbiah, Kaplan, Dhariwal,
  Neelakantan, Shyam, Sastry, Askell, Agarwal, Herbert-Voss, Krueger, Henighan,
  Child, Ramesh, Ziegler, Wu, Winter, Hesse, Chen, Sigler, Litwin, Gray, Chess,
  Clark, Berner, McCandlish, Radford, Sutskever, and
  Amodei}]{brown2020language}
Tom~B. Brown, Benjamin Mann, Nick Ryder, Melanie Subbiah, Jared Kaplan,
  Prafulla Dhariwal, Arvind Neelakantan, Pranav Shyam, Girish Sastry, Amanda
  Askell, Sandhini Agarwal, Ariel Herbert-Voss, Gretchen Krueger, Tom Henighan,
  Rewon Child, Aditya Ramesh, Daniel~M. Ziegler, Jeffrey Wu, Clemens Winter,
  Christopher Hesse, Mark Chen, Eric Sigler, Mateusz Litwin, Scott Gray,
  Benjamin Chess, Jack Clark, Christopher Berner, Sam McCandlish, Alec Radford,
  Ilya Sutskever, and Dario Amodei. 2020.
\newblock \href {https://arxiv.org/abs/2005.14165} {Language models are
  few-shot learners}.
\newblock \emph{Preprint}, arXiv:2005.14165.

\bibitem[{Cohen(1960)}]{cohen1960coefficient}
Jacob Cohen. 1960.
\newblock A coefficient of agreement for nominal scales.
\newblock \emph{Educational and psychological measurement}, 20(1):37--46.

\bibitem[{Delaunay et~al.(2023)Delaunay, Tran, Gonz'alez-Gallardo, Bordea,
  Sid{\`e}re, and Doucet}]{Delaunay2023ACS}
Julien Delaunay, Thi Hong~Hanh Tran, Carlos-Emiliano Gonz'alez-Gallardo,
  Georgeta Bordea, Nicolas Sid{\`e}re, and Antoine Doucet. 2023.
\newblock \href {https://api.semanticscholar.org/CorpusID:263139345} {A
  comprehensive survey of document-level relation extraction (2016-2023)}.
\newblock \emph{ArXiv}, abs/2309.16396.

\bibitem[{Gardner et~al.(2017)Gardner, Grus, Neumann, Tafjord, Dasigi, Liu,
  Peters, Schmitz, and Zettlemoyer}]{Gardner2017AllenNLP}
Matt Gardner, Joel Grus, Mark Neumann, Oyvind Tafjord, Pradeep Dasigi,
  Nelson~F. Liu, Matthew Peters, Michael Schmitz, and Luke~S. Zettlemoyer.
  2017.
\newblock \href {https://arxiv.org/abs/arXiv:1803.07640} {Allennlp: A deep
  semantic natural language processing platform}.

\bibitem[{Jiang et~al.(2023)Jiang, Sablayrolles, Mensch, Bamford, Chaplot,
  de~las Casas, Bressand, Lengyel, Lample, Saulnier, Lavaud, Lachaux, Stock,
  Scao, Lavril, Wang, Lacroix, and Sayed}]{jiang2023mistral}
Albert~Q. Jiang, Alexandre Sablayrolles, Arthur Mensch, Chris Bamford,
  Devendra~Singh Chaplot, Diego de~las Casas, Florian Bressand, Gianna Lengyel,
  Guillaume Lample, Lucile Saulnier, Lélio~Renard Lavaud, Marie-Anne Lachaux,
  Pierre Stock, Teven~Le Scao, Thibaut Lavril, Thomas Wang, Timothée Lacroix,
  and William~El Sayed. 2023.
\newblock \href {https://arxiv.org/abs/2310.06825} {Mistral 7b}.
\newblock \emph{Preprint}, arXiv:2310.06825.

\bibitem[{Landis and Koch(1977)}]{Landis1977TheMO}
J~Richard Landis and Gary~G. Koch. 1977.
\newblock The measurement of observer agreement for categorical data.
\newblock \emph{Biometrics}, 33 1:159--74.

\bibitem[{Levin and Hovav(2005)}]{levin2005argument}
B.~Levin and M.R. Hovav. 2005.
\newblock \href {https://books.google.com/books?id=msi9a50gHVYC}
  {\emph{Argument Realization}}.
\newblock Research Surveys in Linguistics. Cambridge University Press.

\bibitem[{Noriega-Atala et~al.(2018)Noriega-Atala, Hein, Thumsi, Wong, Wang,
  and Morrison}]{NoriegaAtala2018InterSentenceRE}
Enrique Noriega-Atala, Paul~Douglas Hein, Shraddha~Satish Thumsi, Zechy Wong,
  Xia Wang, and Clayton~T. Morrison. 2018.
\newblock \href {https://api.semanticscholar.org/CorpusID:56379048}
  {Inter-sentence relation extraction for associating biological context with
  events in biomedical texts}.
\newblock \emph{2018 IEEE International Conference on Data Mining Workshops
  (ICDMW)}, pages 722--731.

\bibitem[{Noriega{-}Atala et~al.(2021)Noriega{-}Atala, Lovett, Morrison, and
  Surdeanu}]{DBLP:journals/corr/abs-2112-09288}
Enrique Noriega{-}Atala, Peter~M. Lovett, Clayton~T. Morrison, and Mihai
  Surdeanu. 2021.
\newblock \href {https://arxiv.org/abs/2112.09288} {Neural architectures for
  biological inter-sentence relation extraction}.
\newblock \emph{CoRR}, abs/2112.09288.

\bibitem[{OpenAI(2023)}]{openai2023gpt4}
OpenAI. 2023.
\newblock \href {https://arxiv.org/abs/2303.08774} {Gpt-4 technical report}.
\newblock \emph{Preprint}, arXiv:2303.08774.

\bibitem[{Raffel et~al.(2019)Raffel, Shazeer, Roberts, Lee, Narang, Matena,
  Zhou, Li, and Liu}]{Raffel2019ExploringTL}
Colin Raffel, Noam~M. Shazeer, Adam Roberts, Katherine Lee, Sharan Narang,
  Michael Matena, Yanqi Zhou, Wei Li, and Peter~J. Liu. 2019.
\newblock \href {https://api.semanticscholar.org/CorpusID:204838007} {Exploring
  the limits of transfer learning with a unified text-to-text transformer}.
\newblock \emph{J. Mach. Learn. Res.}, 21:140:1--140:67.

\bibitem[{Rajpurkar et~al.(2016)Rajpurkar, Zhang, Lopyrev, and
  Liang}]{Rajpurkar2016SQuAD1Q}
Pranav Rajpurkar, Jian Zhang, Konstantin Lopyrev, and Percy Liang. 2016.
\newblock \href {https://api.semanticscholar.org/CorpusID:11816014} {Squad:
  100,000+ questions for machine comprehension of text}.
\newblock In \emph{Conference on Empirical Methods in Natural Language
  Processing}.

\bibitem[{Sahu et~al.(2019)Sahu, Christopoulou, Miwa, and
  Ananiadou}]{sahu-etal-2019-inter}
Sunil~Kumar Sahu, Fenia Christopoulou, Makoto Miwa, and Sophia Ananiadou. 2019.
\newblock \href {https://doi.org/10.18653/v1/P19-1423} {Inter-sentence relation
  extraction with document-level graph convolutional neural network}.
\newblock In \emph{Proceedings of the 57th Annual Meeting of the Association
  for Computational Linguistics}, pages 4309--4316, Florence, Italy.
  Association for Computational Linguistics.

\bibitem[{Sosa and Altman(2022)}]{Sosa2022ContextsAC}
Daniel~N. Sosa and Russ~B. Altman. 2022.
\newblock \href {https://api.semanticscholar.org/CorpusID:250453021} {Contexts
  and contradictions: a roadmap for computational drug repurposing with
  knowledge inference}.
\newblock \emph{Briefings in Bioinformatics}, 23.

\bibitem[{Sutskever et~al.(2014)Sutskever, Vinyals, and
  Le}]{Sutskever2014SequenceTS}
Ilya Sutskever, Oriol Vinyals, and Quoc~V. Le. 2014.
\newblock \href {https://api.semanticscholar.org/CorpusID:7961699} {Sequence to
  sequence learning with neural networks}.
\newblock \emph{ArXiv}, abs/1409.3215.

\bibitem[{Touvron et~al.(2023)Touvron, Lavril, Izacard, Martinet, Lachaux,
  Lacroix, Rozi{\`e}re, Goyal, Hambro, Azhar, Rodriguez, Joulin, Grave, and
  Lample}]{Touvron2023LLaMAOA}
Hugo Touvron, Thibaut Lavril, Gautier Izacard, Xavier Martinet, Marie-Anne
  Lachaux, Timoth{\'e}e Lacroix, Baptiste Rozi{\`e}re, Naman Goyal, Eric
  Hambro, Faisal Azhar, Aurelien Rodriguez, Armand Joulin, Edouard Grave, and
  Guillaume Lample. 2023.
\newblock \href {https://api.semanticscholar.org/CorpusID:257219404} {Llama:
  Open and efficient foundation language models}.
\newblock \emph{ArXiv}, abs/2302.13971.

\bibitem[{Vaswani et~al.(2017)Vaswani, Shazeer, Parmar, Uszkoreit, Jones,
  Gomez, Kaiser, and Polosukhin}]{Vaswani2017AttentionIA}
Ashish Vaswani, Noam~M. Shazeer, Niki Parmar, Jakob Uszkoreit, Llion Jones,
  Aidan~N. Gomez, Lukasz Kaiser, and Illia Polosukhin. 2017.
\newblock \href {https://api.semanticscholar.org/CorpusID:13756489} {Attention
  is all you need}.
\newblock In \emph{Neural Information Processing Systems}.

\bibitem[{Xu et~al.(2022)Xu, Chen, Mou, and Zhao}]{xu-etal-2022-document}
Wang Xu, Kehai Chen, Lili Mou, and Tiejun Zhao. 2022.
\newblock \href {https://doi.org/10.18653/v1/2022.naacl-main.212}
  {Document-level relation extraction with sentences importance estimation and
  focusing}.
\newblock In \emph{Proceedings of the 2022 Conference of the North American
  Chapter of the Association for Computational Linguistics: Human Language
  Technologies}, pages 2920--2929, Seattle, United States. Association for
  Computational Linguistics.

\end{thebibliography}
